\documentclass[11pt]{article}
\usepackage{acl2016}
\usepackage{times}
\usepackage{latexsym}
\usepackage{url}

\aclfinalcopy 
\def\aclpaperid{1} 

\usepackage{yfonts}
\usepackage{graphicx} 
\usepackage{subfigure}

\usepackage[linesnumbered,ruled,vlined]{algorithm2e}

\SetCommentSty{mycommfont}

\usepackage{verbatim}
\usepackage{etoolbox}
\usepackage{framed}
\usepackage{xcolor}
\usepackage{color}
\usepackage{amsmath}
\newcommand{\argmax}{\operatornamewithlimits{argmax}}

\usepackage{multirow}
\usepackage{arydshln}

\title{End-to-end Sequence Labeling via Bi-directional LSTM-CNNs-CRF}

\author{Xuezhe Ma \and Eduard Hovy \\
	    Language Technologies Institute \\
	    Carnegie Mellon University \\
	    Pittsburgh, PA 15213, USA \\
	    {\tt xuezhem@cs.cmu.edu, ehovy@cmu.edu}}

\date{}

\begin{document}

\maketitle

\begin{abstract}
State-of-the-art sequence labeling systems traditionally require large amounts of task-specific knowledge in the form of hand-crafted features and data pre-processing. In this paper, we introduce a novel neutral network architecture that benefits from both word- and character-level representations automatically, by using combination of bidirectional LSTM, CNN and CRF. Our system is truly end-to-end, requiring no feature engineering or data pre-processing, thus making it applicable to a wide range of sequence labeling tasks. We evaluate our system on two data sets for two sequence labeling tasks --- Penn Treebank WSJ corpus for part-of-speech (POS) tagging and CoNLL 2003 corpus for named entity recognition (NER). We obtain state-of-the-art performance on both datasets --- 97.55\% accuracy for POS tagging and 91.21\% F1 for NER.
\end{abstract}

\section{Introduction}
Linguistic sequence labeling, such as part-of-speech (POS) tagging and named entity recognition (NER), is one of the first stages in deep language understanding and its importance has been well recognized in the natural language processing community. Natural language processing (NLP) systems, like syntactic parsing~\cite{Nivre:2004,McDonald:2005,Koo:2010,ma-zhao:2012:POSTERS,ma2015probabilistic,chen-manning:2014:EMNLP2014,ma-hovy:2015:EMNLP} and entity coreference resolution~\cite{ng:2010:ACL,ma-hovy:2016:NAACL}, are becoming more sophisticated, in part because of utilizing output information of POS tagging or NER systems.

Most traditional high performance sequence labeling models are linear statistical models, including Hidden Markov Models (HMM) and Conditional Random Fields (CRF)~\cite{ratinov2009design,passos-kumar-mccallum:2014:W14-16,luo-EtAl:2015:EMNLP2}, which rely heavily on hand-crafted features and task-specific resources. For example, English POS taggers benefit from carefully designed word spelling features; orthographic features and external resources such as gazetteers are widely used in NER. However, such task-specific knowledge is costly to develop~\cite{ma-xia:2014:P14-1}, making sequence labeling models difficult to adapt to new tasks or new domains.

In the past few years, non-linear neural networks with as input distributed word representations, also known as word embeddings, have been broadly applied to NLP problems with great success. \newcite{collobert2011natural} proposed a simple but effective feed-forward neutral network that independently classifies labels for each word by using contexts within a window with fixed size. Recently, recurrent neural networks (RNN)~\cite{goller1996learning}, together with its variants such as long-short term memory (LSTM)~\cite{hochreiter1997long,gers2000learning} and gated recurrent unit (GRU)~\cite{cho2014properties}, have shown great success in modeling sequential data.
Several RNN-based neural network models have been proposed to solve sequence labeling tasks like speech recognition~\cite{graves2013speech}, POS tagging~\cite{huang2015bidirectional} and NER~\cite{chiu2015named,Hu:2016:ACL}, achieving competitive performance against traditional models. However, even systems that have utilized distributed representations as inputs have used these to augment, rather than replace, hand-crafted features (e.g. word spelling and capitalization patterns). Their performance drops rapidly when the models solely depend on neural embeddings.

In this paper, we propose a neural network architecture for sequence labeling. It is a truly end-to-end model requiring no task-specific resources, feature engineering, or data pre-processing beyond pre-trained word embeddings on unlabeled corpora. Thus, our model can be easily applied to a wide range of sequence labeling tasks on different languages and domains. We first use convolutional neural networks (CNNs)~\cite{lecun1989backpropagation} to encode character-level information of a word into its character-level representation. Then we combine character- and word-level representations and feed them into bi-directional LSTM (BLSTM) to model context information of each word. On top of BLSTM, we use a sequential CRF to jointly decode labels for the whole sentence. We evaluate our model on two linguistic sequence labeling tasks --- POS tagging on Penn Treebank WSJ~\cite{Marcus:1993}, and NER on English data from the CoNLL 2003 shared task~\cite{TjongKimSang:2003}. Our end-to-end model outperforms previous state-of-the-art systems, obtaining 97.55\% accuracy for POS tagging and 91.21\% F1 for NER. The contributions of this work are (i) proposing a novel neural network architecture for linguistic sequence labeling. (ii) giving empirical evaluations of this model on benchmark data sets for two classic NLP tasks. (iii) achieving state-of-the-art performance with this truly end-to-end system.

\section{Neural Network Architecture}
\label{sec:architec}
In this section, we describe the components (layers) of our neural network architecture.
We introduce the neural layers in our neural network one-by-one from bottom to top.

\subsection{CNN for Character-level Representation}
Previous studies~\cite{santos2014learning,chiu2015named} have shown that CNN is an effective approach to extract morphological information (like the prefix or suffix of a word) from characters of words and encode it into neural representations. Figure~\ref{fig:cnn} shows the CNN we use to extract character-level representation of a given word. The CNN is similar to the one in \newcite{chiu2015named}, except that we use only character embeddings as the inputs to CNN, without character type features. A dropout layer~\cite{srivastava2014dropout} is applied before character embeddings are input to CNN.

\subsection{Bi-directional LSTM}
\subsubsection{LSTM Unit}
Recurrent neural networks (RNNs) are a powerful family of connectionist models that capture time
dynamics via cycles in the graph. Though, in theory, RNNs are capable to capturing long-distance dependencies, 
in practice, they fail due to the gradient vanishing/exploding problems~\cite{bengio1994,pascanu2012}.

\begin{figure}[t]
\centering
\includegraphics[scale=0.72]{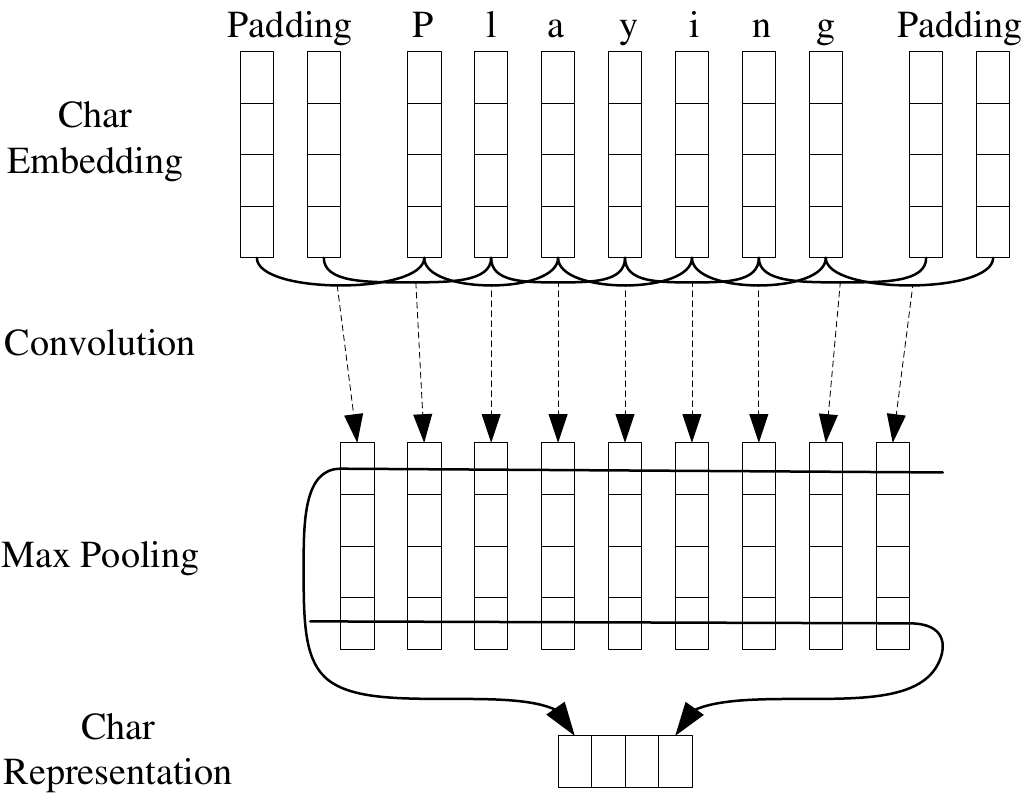}
\caption{The convolution neural network for extracting character-level representations of words. Dashed arrows indicate a dropout layer applied before character embeddings are input to CNN.}
\label{fig:cnn}
\end{figure}

\begin{figure}[h]
\centering
\includegraphics[scale=0.65]{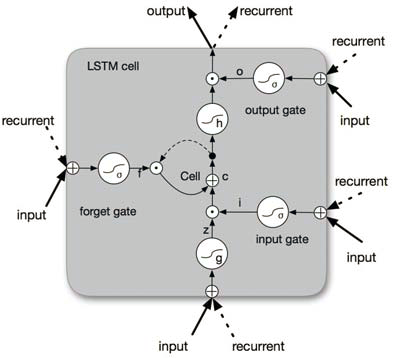}
\caption{Schematic of LSTM unit.}
\label{fig:lstm}
\end{figure}

LSTMs~\cite{hochreiter1997long} are variants of RNNs designed to cope with these gradient vanishing problems. Basically, a LSTM unit is composed of three multiplicative gates which control the proportions of information to forget and to pass on to the next time step. Figure~\ref{fig:lstm} gives the basic structure of an LSTM unit.

Formally, the formulas to update an LSTM unit at time $t$ are:
\begin{displaymath}
\begin{array}{rcl}
\mathbf{i}_{t} & = & \sigma(\boldsymbol{W}_i \mathbf{h}_{t-1} + \boldsymbol{U}_i \mathbf{x}_t + \boldsymbol{b}_i) \\
\mathbf{f}_{t} & = & \sigma(\boldsymbol{W}_f \mathbf{h}_{t-1} + \boldsymbol{U}_f \mathbf{x}_t + \boldsymbol{b}_f) \\
\mathbf{\tilde{c}}_t & = & \mathrm{tanh}(\boldsymbol{W}_c \mathbf{h}_{t-1} + \boldsymbol{U}_c \mathbf{x}_t + \boldsymbol{b}_c) \\
\mathbf{c}_{t} & = & \mathbf{f}_t \odot \mathbf{c}_{t-1} + \mathbf{i}_t \odot \mathbf{\tilde{c}}_t \\
\mathbf{o}_{t} & = & \sigma(\boldsymbol{W}_o \mathbf{h}_{t-1} + \boldsymbol{U}_o \mathbf{x}_t + \boldsymbol{b}_o) \\
\mathbf{h}_{t} & = & \mathbf{o}_{t} \odot \mathrm{tanh}(\mathbf{c}_t)
\end{array}
\end{displaymath}
where $\sigma$ is the element-wise sigmoid function and $\odot$ is the element-wise product. $\mathbf{x}_t$ is the input vector (e.g. word embedding) at time $t$, and $\mathbf{h}_t$ is the hidden state (also called output) vector storing all the useful information at (and before) time $t$. 
$\boldsymbol{U}_i, \boldsymbol{U}_f, \boldsymbol{U}_c, \boldsymbol{U}_o$ denote the weight matrices of different gates for input $\mathbf{x}_t$, and $\boldsymbol{W}_i, \boldsymbol{W}_f, \boldsymbol{W}_c, \boldsymbol{W}_o$ are the weight matrices for hidden state $\mathbf{h}_t$. $\boldsymbol{b}_i, \boldsymbol{b}_f, \boldsymbol{b}_c, \boldsymbol{b}_o$ denote the bias vectors. It should be noted that we do not include peephole connections~\cite{gers2003learning} in the our LSTM formulation.

\subsubsection{BLSTM}
For many sequence labeling tasks it is beneficial to have access to both past (left) and future (right) contexts. However, the LSTM's hidden state $\mathbf{h}_t$ takes information only from past, knowing nothing about the future. An elegant solution whose effectiveness has been proven by previous work~\cite{dyer-EtAl:2015:ACL-IJCNLP} is bi-directional LSTM (BLSTM). The basic idea is to present each sequence forwards and backwards to two separate hidden states to capture past and future information, respectively. Then the two hidden states are concatenated to form the final output.

\subsection{CRF}
For sequence labeling (or general structured prediction) tasks, it is beneficial to consider the correlations between labels in neighborhoods and jointly decode the best chain of labels for a given input sentence. For example, in POS tagging an adjective is more likely to be followed by a noun than a verb, and in NER with standard \textsf{BIO2} annotation~\cite{tksveenstra99eacl} \mbox{\textup{I-ORG}} cannot follow \textup{I-PER}. Therefore, we model label sequence jointly using a conditional random field (CRF)~\cite{lafferty2001}, instead of decoding each label independently.

Formally, we use $\mathbf{z} = \{\mathbf{z}_1, \cdots, \mathbf{z}_n\}$ to represent a generic input sequence where $\mathbf{z}_i$ is the input vector of the $i$th word. 
$\boldsymbol{y} = \{y_1, \cdots, y_n\}$ represents a generic sequence of labels for $\mathbf{z}$.
$\mathcal{Y}(\mathbf{z})$ denotes the set of possible label sequences for $\mathbf{z}$.
The probabilistic model for sequence CRF defines a family of conditional probability $p(\boldsymbol{y}|\mathbf{z};\mathbf{W}, \mathbf{b})$ over all possible label sequences $\boldsymbol{y}$ given $\mathbf{z}$ with the following form:
\begin{displaymath}
p(\boldsymbol{y}|\mathbf{z};\mathbf{W},\mathbf{b}) = \frac{\prod\limits_{i=1}^{n} \psi_i(y_{i-1}, y_i, \mathbf{z})}{\sum\limits_{y' \in \mathcal{Y}(\mathbf{z})} \prod\limits_{i=1}^{n} \psi_i(y'_{i-1}, y'_i, \mathbf{z})}
\end{displaymath}
where $\psi_i(y', y, \mathbf{z}) = \exp(\mathbf{W}_{y',y}^{T} \mathbf{z}_i + \mathbf{b}_{y',y})$ are potential functions, and $\mathbf{W}_{y',y}^{T}$ and $\mathbf{b}_{y',y}$ are the weight vector and bias corresponding to label pair $(y', y)$, respectively.

For CRF training, we use the maximum conditional likelihood estimation. For a training set 
$\{(\mathbf{z}_i, \boldsymbol{y}_i)\}$, the logarithm of the likelihood (a.k.a. the log-likelihood) 
is given by:
\begin{displaymath}
L(\mathbf{W}, \mathbf{b}) = \sum\limits_i \log p(\boldsymbol{y}|\mathbf{z};\mathbf{W}, \mathbf{b})
\end{displaymath}
Maximum likelihood training chooses parameters such that the log-likelihood $L(\mathbf{W}, \mathbf{b})$ is maximized.

Decoding is to search for the label sequence $\boldsymbol{y}^*$ with the highest conditional probability:
\begin{displaymath}
\boldsymbol{y}^* = \argmax\limits_{y \in \mathcal{Y}(\mathbf{z})} p(\boldsymbol{y}|\mathbf{z};\mathbf{W}, \mathbf{b})
\end{displaymath}
For a sequence CRF model (only interactions between two successive labels are considered), training and decoding can be solved efficiently by adopting the Viterbi algorithm.

\subsection{BLSTM-CNNs-CRF}
\begin{figure}[t]
\centering
\includegraphics[scale=0.9]{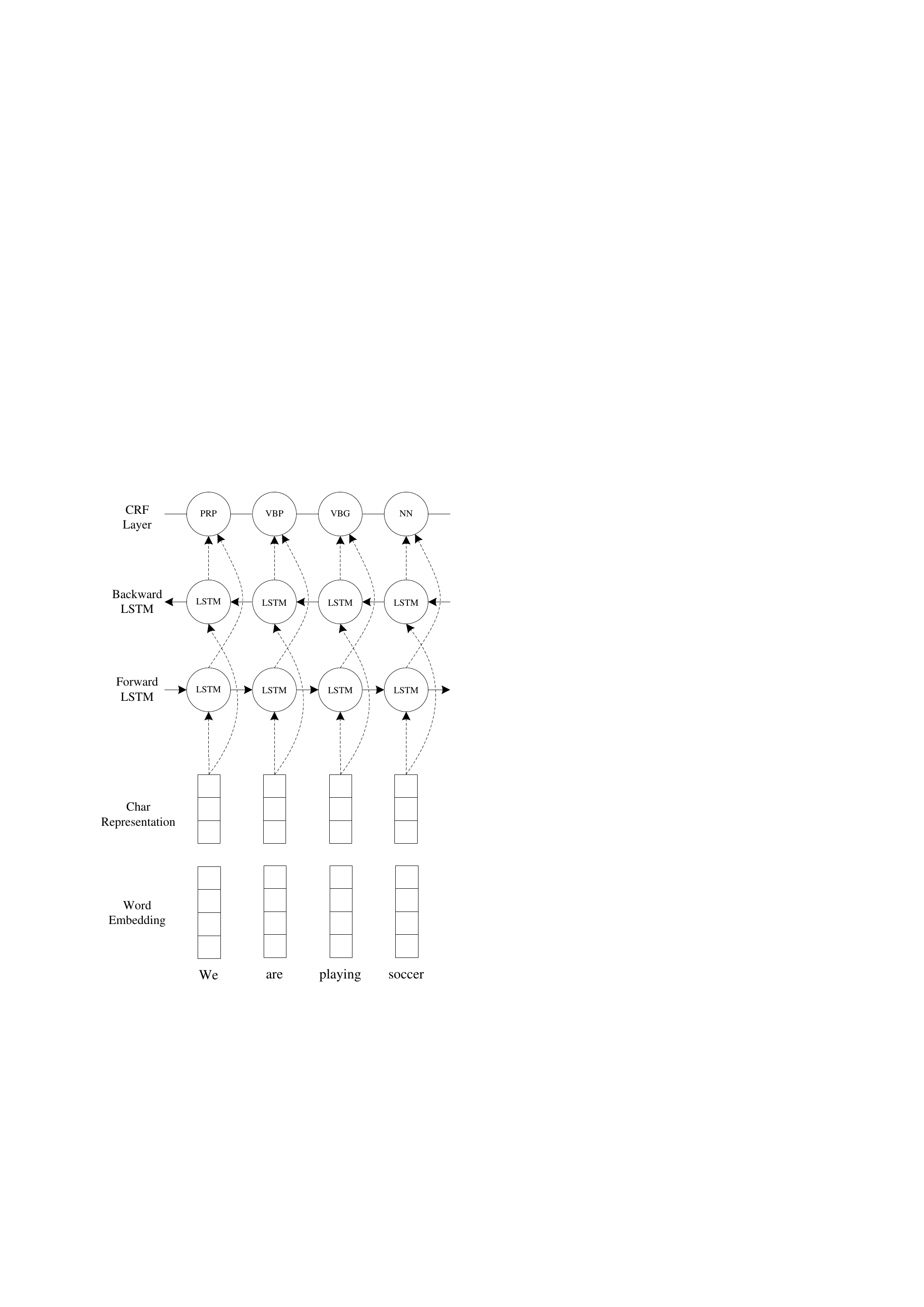}
\caption{The main architecture of our neural network. The character representation for each word is computed by the CNN in Figure~\ref{fig:cnn}. Then the character representation vector is concatenated with the word embedding before feeding into the BLSTM network. Dashed arrows indicate dropout layers applied on both the input and output vectors of BLSTM.}
\label{fig:architec}
\end{figure}

Finally, we construct our neural network model by feeding the output vectors of BLSTM into a CRF layer. Figure~\ref{fig:architec} illustrates the architecture of our network in detail.

For each word, the character-level representation is computed by the CNN in Figure~\ref{fig:cnn} with character embeddings as inputs. Then the character-level representation vector is concatenated with the word embedding vector to feed into the BLSTM network. Finally, the output vectors of BLSTM are fed to the CRF layer to jointly decode the best label sequence. As shown in Figure~\ref{fig:architec}, dropout layers are applied on both the input and output vectors of BLSTM. Experimental results show that using dropout significantly improve the performance of our model (see Section~\ref{subsec:dropout} for details).

\section{Network Training}
In this section, we provide details about training the neural network. We implement the neural network using the Theano library~\cite{bergstra2010theano}. The computations for a single model are run on a GeForce GTX TITAN X GPU. Using the settings discussed in this section, the model training requires about 12 hours for POS tagging and 8 hours for NER.

\subsection{Parameter Initialization}
\label{subsec:init}
\textbf{Word Embeddings.} We use Stanford's publicly available GloVe 100-dimensional embeddings\footnote{\url{http://nlp.stanford.edu/projects/glove/}} trained on 6 billion words from Wikipedia and web text~\cite{pennington:EMNLP2014}

We also run experiments on two other sets of published embeddings, namely Senna 50-dimensional embeddings\footnote{\url{http://ronan.collobert.com/senna/}} trained on Wikipedia and Reuters RCV-1 corpus~\cite{collobert2011natural}, and Google's Word2Vec 300-dimensional embeddings\footnote{\url{https://code.google.com/archive/p/word2vec/}} trained on 100 billion words from Google News~\cite{mikolov2013}. To test the effectiveness of pretrained word embeddings, we experimented with randomly initialized embeddings with 100 dimensions, where embeddings are uniformly sampled from range $[-\sqrt{\frac{3}{dim}}, +\sqrt{\frac{3}{dim}}]$ where $dim$ is the dimension of embeddings~\cite{he2015delving}. The performance of different word embeddings is discussed in Section~\ref{subsec:embedding}.

\noindent
\textbf{Character Embeddings.} Character embeddings are initialized with uniform samples from $[-\sqrt{\frac{3}{dim}}, +\sqrt{\frac{3}{dim}}]$, where we set $dim=30$.

\noindent
\textbf{Weight Matrices and Bias Vectors.} Matrix parameters are randomly initialized with uniform samples from $[-\sqrt{\frac{6}{r + c}}, +\sqrt{\frac{6}{r + c}}]$, where $r$ and $c$ are the number of of rows and columns in the structure~\cite{glorot2010}. Bias vectors are initialized to zero, except the bias $\mathbf{b}_f$ for the forget gate in LSTM , which is initialized to 1.0~\cite{jozefowicz2015}.

\subsection{Optimization Algorithm}
Parameter optimization is performed with mini-batch stochastic gradient descent (SGD) with batch size 10 and momentum 0.9.
We choose an initial learning rate of $\eta_0$ ($\eta_0 = 0.01$ for POS tagging, and $0.015$ for NER, see Section~\ref{subsec:hyper-params}.), and the learning rate is updated on each epoch of training as $\eta_t = \eta_0/(1 + \rho t)$, with decay rate $\rho = 0.05$ and $t$ is the number of epoch completed. To reduce the effects of ``gradient exploding'', we use a gradient clipping of $5.0$~\cite{pascanu2012}. We explored other more sophisticated optimization algorithms such as AdaDelta~\cite{zeiler2012adadelta}, Adam~\cite{kingma2014adam} or RMSProp~\cite{dauphin2015rmsprop}, but none of them meaningfully improve upon SGD with momentum and gradient clipping in our preliminary experiments.

\noindent
\textbf{Early Stopping.} We use early stopping~\cite{giles2001,graves2013speech} based on performance on validation sets. The ``best'' parameters appear at around 50 epochs, according to our experiments.

\noindent
\textbf{Fine Tuning.} For each of the embeddings, we fine-tune initial embeddings, modifying them during gradient updates of the neural network model by back-propagating gradients. The effectiveness of this method has been previously explored in sequential and structured prediction problems~\cite{collobert2011natural,peng-dredze:2015:EMNLP}.

\noindent
\textbf{Dropout Training.} To mitigate overfitting, we apply the dropout method~\cite{srivastava2014dropout} to regularize our model. As shown in Figure~\ref{fig:cnn} and \ref{fig:architec}, we apply dropout on character embeddings before inputting to CNN, and on both the input and output vectors of BLSTM. We fix dropout rate at $0.5$ for all dropout layers through all the experiments. We obtain significant improvements on model performance after using dropout (see Section~\ref{subsec:dropout}).

\begin{table}
\centering
\begin{tabular}[t]{l|l|r|r}
\hline
\textbf{Layer} & \textbf{Hyper-parameter} & \textbf{POS} & \textbf{NER} \\
\hline
\multirow{2}{*}{CNN} & window size & 3 & 3 \\
 & number of filters & 30 & 30 \\
\hline
\multirow{3}{*}{LSTM} & state size & 200 & 200 \\
 & initial state & 0.0 & 0.0 \\
 & peepholes & no & no \\
\hline
Dropout & dropout rate & 0.5 & 0.5 \\
\hline
 & batch size & 10 & 10 \\
 & initial learning rate & 0.01 & 0.015 \\
 & decay rate & 0.05 & 0.05 \\
 & gradient clipping & 5.0 & 5.0 \\
\hline
\end{tabular}
\caption{Hyper-parameters for all experiments.}
\label{tab:hyper-params}
\end{table}

\subsection{Tuning Hyper-Parameters}
\label{subsec:hyper-params}
Table~\ref{tab:hyper-params} summarizes the chosen hyper-parameters for all experiments. We tune the hyper-parameters on the development sets by random search. Due to time constrains it is infeasible to do a random search across the full hyper-parameter space. Thus, for the tasks of POS tagging and NER we try to share as many hyper-parameters as possible. Note that the final hyper-parameters for these two tasks are almost the same, except the initial learning rate. We set the state size of LSTM to $200$. Tuning this parameter did not significantly impact the performance of our model. For CNN, we use 30 filters with window length 3. 

\section{Experiments}
\subsection{Data Sets}
As mentioned before, we evaluate our neural network model on two sequence labeling tasks: POS tagging and NER.

\noindent
\textbf{POS Tagging}. For English POS tagging, we use the Wall Street Journal (WSJ) portion of Penn Treebank (PTB)~\cite{Marcus:1993}, which contains 45 different POS tags. In order to compare with previous work, we adopt the standard splits --- section 0--18 as training data, section 19--21 as development data and section 22--24 as test data~\cite{manning2011part,sogaard:2011:ACL-HLT20111}.

\noindent
\textbf{NER}. For NER, We perform experiments on the English data from CoNLL 2003 shared task~\cite{TjongKimSang:2003}. This data set contains four different types of named entities: \emph{PERSON, LOCATION, ORGANIZATION}, and \emph{MISC}. We use the \textsf{BIOES} tagging scheme instead of standard \textsf{BIO2}, as previous studies have reported meaningful improvement with this scheme~\cite{ratinov2009design,dai2015enhancing,lample:2016:NAACL}.

The corpora statistics are shown in Table~\ref{tab:corpus}. We did not perform any pre-processing for data sets, leaving our system truly end-to-end.

\begin{table}
\centering
\begin{tabular}[t]{l|c|c|c}
\hline
\textbf{Dataset} & & \textbf{WSJ} & \textbf{CoNLL2003} \\
\hline
\multirow{2}{*}{Train} & SENT & 38,219 & 14,987\\
 & TOKEN & 912,344 & 204,567 \\
\hline
\multirow{2}{*}{Dev} & SENT & 5,527 & 3,466 \\
 & TOKEN & 131,768 & 51,578 \\
\hline
\multirow{2}{*}{Test} & SENT & 5,462 & 3,684 \\
 & TOKEN & 129,654 & 46,666 \\
\hline
\end{tabular}
\caption{Corpora statistics. SENT and TOKEN refer to the number of sentences and tokens in each data set.}
\label{tab:corpus}
\end{table}

\begin{table*}
\centering
\begin{tabular}[t]{l|cc|ccc:ccc}
\hline
 & \multicolumn{2}{c|}{\textbf{POS}} & \multicolumn{6}{c}{\textbf{NER}} \\
 \cline{2-9}
 & \textbf{Dev} & \textbf{Test} & \multicolumn{3}{c:}{\textbf{Dev}} & \multicolumn{3}{c}{\textbf{Test}} \\
 \cline{2-9}
\textbf{Model} & Acc. & Acc. & Prec. & Recall & F1 & Prec. & Recall & F1 \\
\hline
BRNN & 96.56 & 96.76 & 92.04 & 89.13 & 90.56 & 87.05 & 83.88 & 85.44 \\
BLSTM & 96.88 & 96.93 & 92.31 & 90.85 & 91.57 & 87.77 & 86.23 & 87.00 \\
BLSTM-CNN & 97.34 & 97.33 & 92.52 & 93.64 & 93.07 & 88.53 & 90.21 & 89.36 \\
BRNN-CNN-CRF & 97.46 & 97.55 & 94.85 & 94.63 & 94.74 & 91.35 & 91.06 & 91.21 \\
\hline
\end{tabular}
\caption{Performance of our model on both the development and test sets of the two tasks, together with three baseline systems.}
\label{tab:result}
\end{table*}

\subsection{Main Results}
\label{subsec:results}
We first run experiments to dissect the effectiveness of each component (layer) of our neural network architecture by ablation studies. We compare the performance with three baseline systems --- BRNN, the bi-direction RNN; BLSTM, the bi-direction LSTM, and BLSTM-CNNs, the combination of BLSTM with CNN to model character-level information. All these models are run using Stanford's GloVe 100 dimensional word embeddings and the same hyper-parameters as shown in Table~\ref{tab:hyper-params}. According to the results shown in Table~\ref{tab:result}, BLSTM obtains better performance than BRNN on all evaluation metrics of both the two tasks. BLSTM-CNN models significantly outperform the BLSTM model, showing that character-level representations are important for linguistic sequence labeling tasks. This is consistent with results reported by previous work~\cite{santos2014learning,chiu2015named}. Finally, by adding CRF layer for joint decoding we achieve significant improvements over BLSTM-CNN models for both POS tagging and NER on all metrics. This demonstrates that jointly decoding label sequences can significantly benefit the final performance of neural network models.

\begin{table}
\centering
\begin{tabular}[t]{l|c}
\hline
\textbf{Model} & \textbf{Acc.} \\
\hline
\newcite{gimenez2004svmtool} & 97.16 \\
\newcite{toutanova2003feature} & 97.27 \\ 
\newcite{manning2011part} & 97.28 \\
\newcite{collobert2011natural}$^\ddag$ & 97.29 \\
\newcite{santos2014learning}$^\ddag$ & 97.32 \\
\newcite{shen2007guided} & 97.33 \\
\newcite{sun2014structure} & 97.36 \\
\newcite{sogaard:2011:ACL-HLT20111} & 97.50 \\
\hline
\textbf{This paper} & \textbf{97.55} \\
\hline
\end{tabular}
\caption{POS tagging accuracy of our model on test data from WSJ proportion of PTB, together with top-performance systems. The neural network based models are marked with $\ddag$.}
\label{tab:pos}
\end{table}

\subsection{Comparison with Previous Work}
\label{subsec:compare}
\subsubsection{POS Tagging}
Table~\ref{tab:pos} illustrates the results of our model for POS tagging, together with seven previous top-performance systems for comparison. Our model significantly outperform Senna~\cite{collobert2011natural}, which is a feed-forward neural network model using capitalization and discrete suffix features, and data pre-processing. Moreover, our model achieves 0.23\% improvements on accuracy over the ``CharWNN''~\cite{santos2014learning}, which is a neural network model based on Senna and also uses CNNs to model character-level representations. This demonstrates the effectiveness of BLSTM for modeling sequential data and the importance of joint decoding with structured prediction model. 

Comparing with traditional statistical models, our system achieves state-of-the-art accuracy, obtaining 0.05\% improvement over the previously best reported results by \newcite{sogaard:2011:ACL-HLT20111}. It should be noted that \newcite{huang2015bidirectional} also evaluated their BLSTM-CRF model for POS tagging on WSJ corpus. But they used a different splitting of the training/dev/test data sets. Thus, their results are not directly comparable with ours.

\begin{table}
\centering
\begin{tabular}[t]{l|c}
\hline
\textbf{Model} & \textbf{F1} \\
\hline
\newcite{chieu2002named} & 88.31 \\
\newcite{florian2003named} & 88.76 \\
\newcite{ando2005framework} & 89.31 \\
\newcite{collobert2011natural}$^\ddag$ & 89.59 \\
\newcite{huang2015bidirectional}$^\ddag$ & 90.10 \\
\newcite{chiu2015named}$^\ddag$ & 90.77 \\
\newcite{ratinov2009design} & 90.80 \\
\newcite{lin2009phrase} & 90.90 \\
\newcite{passos-kumar-mccallum:2014:W14-16} & 90.90 \\
\newcite{lample:2016:NAACL}$^\ddag$ & 90.94 \\
\newcite{luo-EtAl:2015:EMNLP2} & 91.20 \\
\hline
\textbf{This paper} & \textbf{91.21} \\
\hline
\end{tabular}
\caption{NER F1 score of our model on test data set from CoNLL-2003. For the purpose of comparison, we also list F1 scores of previous top-performance systems. $\ddag$ marks the neural models.}
\label{tab:ner}
\end{table}

\subsubsection{NER}
Table~\ref{tab:ner} shows the F1 scores of previous models for NER on the test data set from CoNLL-2003 shared task. For the purpose of comparison, we list their results together with ours. Similar to the observations of POS tagging, our model achieves significant improvements over Senna and the other three neural models, namely the LSTM-CRF proposed by \newcite{huang2015bidirectional}, LSTM-CNNs proposed by \newcite{chiu2015named}, and the LSTM-CRF by \newcite{lample:2016:NAACL}. \newcite{huang2015bidirectional} utilized discrete spelling, POS and context features, \newcite{chiu2015named} used character-type, capitalization, and lexicon features, and all the three model used some task-specific data pre-processing, while our model does not require any carefully designed features or data pre-processing. We have to point out that the result (90.77\%) reported by \newcite{chiu2015named} is incomparable with ours, because their final model was trained on the combination of the training and development data sets\footnote{We run experiments using the same setting and get 91.37\% F1 score.}. 

To our knowledge, the previous best F1 score (91.20)\footnote{Numbers are taken from the Table 3 of the original paper~\cite{luo-EtAl:2015:EMNLP2}. While there is clearly inconsistency among the precision (91.5\%), recall (91.4\%) and F1 scores (91.2\%), it is unclear in which way they are incorrect.} reported on CoNLL 2003 data set is by the joint NER and entity linking model~\cite{luo-EtAl:2015:EMNLP2}. This model used many hand-crafted features including stemming and spelling features, POS and chunks tags, WordNet clusters, Brown Clusters, as well as external knowledge bases such as Freebase and Wikipedia. Our end-to-end model slightly improves this model by 0.01\%, yielding a state-of-the-art performance.

\begin{table}
\centering
\begin{tabular}[t]{l|c|c|c}
\hline
\textbf{Embedding} & Dimension & POS & NER \\
\hline
Random & 100 & 97.13 & 80.76 \\
Senna & 50 & 97.44 & 90.28 \\
Word2Vec & 300 & 97.40 & 84.91 \\
GloVe & 100 & \textbf{97.55} & \textbf{91.21} \\
\hline
\end{tabular}
\caption{Results with different choices of word embeddings on the two 
tasks (accuracy for POS tagging and F1 for NER).}
\label{tab:embedding}
\end{table}

\subsection{Word Embeddings}
\label{subsec:embedding}
As mentioned in Section~\ref{subsec:init}, in order to test the importance of pretrained word embeddings, we performed experiments with different sets of publicly published word embeddings, as well as a random sampling method, to initialize our model. Table~\ref{tab:embedding} gives the performance of three different word embeddings, as well as the randomly sampled one. According to the results in Table~\ref{tab:embedding}, models using pretrained word embeddings obtain a significant improvement as opposed to the ones using random embeddings. Comparing the two tasks, NER relies more heavily on pretrained embeddings than POS tagging. This is consistent with results reported by previous work~\cite{collobert2011natural,huang2015bidirectional,chiu2015named}.

For different pretrained embeddings, Stanford's GloVe 100 dimensional embeddings achieve best results on both tasks, about 0.1\% better on POS accuracy and 0.9\% better on NER F1 score than the Senna 50 dimensional one. This is different from the results reported by \newcite{chiu2015named}, where Senna achieved slightly better performance on NER than other embeddings. Google's Word2Vec 300 dimensional embeddings obtain similar performance with Senna on POS tagging, still slightly behind GloVe. But for NER, the performance on Word2Vec is far behind GloVe and Senna. One possible reason that Word2Vec is not as good as the other two embeddings on NER is because of vocabulary mismatch --- Word2Vec embeddings were trained in case-sensitive manner, excluding many common symbols such as punctuations and digits. Since we do not use any data pre-processing to deal with such common symbols or rare words, it might be an issue for using Word2Vec.

\begin{table}
\centering
{\small
\begin{tabular}[t]{l|ccc|ccc}
\hline
 & \multicolumn{3}{c|}{\textbf{POS}} & \multicolumn{3}{c}{\textbf{NER}} \\
 \cline{2-7}
 & \textbf{Train} & \textbf{Dev} & \textbf{Test} & \textbf{Train} & \textbf{Dev} & \textbf{Test} \\
\hline
No & 98.46 & 97.06 & 97.11 & 99.97 & 93.51 & 89.25 \\
Yes & 97.86 & 97.46 & 97.55 & 99.63 & 94.74 & 91.21 \\
\hline
\end{tabular}
}
\caption{Results with and without dropout on two tasks (accuracy for POS tagging and F1 for NER).}
\label{tab:dropout}
\end{table}

\begin{table}
\centering
\begin{tabular}[t]{l|cc|cc}
\hline
 & \multicolumn{2}{c|}{\textbf{POS}} & \multicolumn{2}{c}{\textbf{NER}} \\
 & Dev & Test & Dev & Test \\
\hline
IV & 127,247 & 125,826 & 4,616 & 3,773 \\
OOTV & 2,960 & 2,412 & 1,087 & 1,597 \\
OOEV & 659 & 588 & 44 & 8 \\
OOBV & 902 & 828 & 195 & 270 \\
\hline
\end{tabular}
\caption{Statistics of the partition on each corpus. It lists the number of tokens of each subset for POS tagging and the number of entities for NER.}
\label{tab:part:oov}
\end{table}

\begin{table*}
\centering
\begin{tabular}[t]{l|cccc|cccc}
\hline
 & \multicolumn{8}{c}{\textbf{POS}} \\
 \cline{2-9}
 & \multicolumn{4}{c|}{Dev} & \multicolumn{4}{c}{Test} \\
 \cline{2-9}
 & IV & OOTV & OOEV & OOBV & IV & OOTV & OOEV & OOBV \\
\hline
LSTM-CNN & 97.57 & \textbf{93.75} & 90.29 & 80.27 & 97.55 & \textbf{93.45} & 90.14 & 80.07 \\
LSTM-CNN-CRF & \textbf{97.68} & 93.65 & \textbf{91.05} & \textbf{82.71} & \textbf{97.77} & 93.16 & \textbf{90.65} & \textbf{82.49} \\
\hline
& \multicolumn{8}{c}{\textbf{NER}} \\
 \cline{2-9}
 & \multicolumn{4}{c|}{Dev} & \multicolumn{4}{c}{Test} \\
 \cline{2-9}
 & IV & OOTV & OOEV & OOBV & IV & OOTV & OOEV & OOBV \\
\hline
LSTM-CNN & 94.83 & 87.28 & 96.55 & 82.90 & 90.07 & 89.45 & 100.00 & 78.44 \\
LSTM-CNN-CRF & \textbf{96.49} & \textbf{88.63} & \textbf{97.67} & \textbf{86.91} & \textbf{92.14} & \textbf{90.73} & 100.00 & \textbf{80.60} \\
\hline
\end{tabular}
\caption{Comparison of performance on different subsets of words (accuracy for POS and F1 for NER).}
\label{tab:oov}
\end{table*}

\subsection{Effect of Dropout}
\label{subsec:dropout}
Table~\ref{tab:dropout} compares the results with and without dropout layers for each data set.
All other hyper-parameters remain the same as in Table~\ref{tab:hyper-params}. We observe a essential improvement for both the two tasks. It demonstrates the effectiveness of dropout in reducing overfitting.

\subsection{OOV Error Analysis}
To better understand the behavior of our model, we perform error analysis on Out-of-Vocabulary words (OOV). Specifically, we partition each data set into four subsets --- in-vocabulary words (IV), out-of-training-vocabulary words (OOTV), out-of-embedding-vocabulary words (OOEV) and out-of-both-vocabulary words (OOBV). A word is considered IV if it appears in both the training and embedding vocabulary, while OOBV if neither. OOTV words are the ones do not appear in training set but in embedding vocabulary, while OOEV are the ones do not appear in embedding vocabulary but in training set. For NER, an entity is considered as OOBV if there exists at lease one word not in training set and at least one word not in embedding vocabulary, and the other three subsets can be done in similar manner. Table~\ref{tab:part:oov}
informs the statistics of the partition on each corpus. The embedding we used is Stanford's GloVe with dimension 100, the same as Section~\ref{subsec:results}.

Table~\ref{tab:oov} illustrates the performance of our model on different subsets of words, together with the baseline LSTM-CNN model for comparison. The largest improvements appear on the OOBV subsets of both the two corpora. This demonstrates that by adding CRF for joint decoding, our model is more powerful on words that are out of both the training and embedding sets.

\section{Related Work}
In recent years, several different neural network architectures have been proposed and successfully applied to linguistic sequence labeling such as POS tagging, chunking and NER. Among these neural architectures, the three approaches most similar to our model are the BLSTM-CRF model proposed by \newcite{huang2015bidirectional}, the LSTM-CNNs model by \newcite{chiu2015named} and the BLSTM-CRF by \newcite{lample:2016:NAACL}.

\newcite{huang2015bidirectional} used BLSTM for word-level representations and CRF for jointly label decoding, which is similar to our model. But there are two main differences between their model and ours. First, they did not employ CNNs to model character-level information. Second, they combined their neural network model with hand-crafted features to improve their performance, making their model not an end-to-end system. \newcite{chiu2015named} proposed a hybrid of BLSTM and CNNs to model both character- and word-level representations, which is similar to the first two layers in our model. They evaluated their model on NER and achieved competitive performance. Our model mainly differ from this model by using CRF for joint decoding. Moreover, their model is not truly end-to-end, either, as it utilizes external knowledge such as character-type, capitalization and lexicon features,  and some data pre-processing specifically for NER (e.g. replacing all sequences of digits 0-9 with a single ``0''). Recently, \newcite{lample:2016:NAACL} proposed a BLSTM-CRF model for NER, which utilized BLSTM to model both the character- and word-level information, and use data pre-processing the same as \newcite{chiu2015named}. Instead, we use CNN to model character-level information, achieving better NER performance without using any data pre-processing.

There are several other neural networks previously proposed for sequence labeling. \newcite{labeau2015non} proposed a RNN-CNNs model for German POS tagging. This model is similar to the LSTM-CNNs model in \newcite{chiu2015named}, with the difference of using vanila RNN instead of LSTM. Another neural architecture employing CNN to model character-level information is the ``CharWNN'' architecture~\cite{santos2014learning} which is inspired by the feed-forward network~\cite{collobert2011natural}. CharWNN obtained near state-of-the-art accuracy on English POS tagging (see Section~\ref{subsec:compare} for details). A similar model has also been applied to Spanish and Portuguese NER~\cite{dos2015boosting} \newcite{ling-EtAl:2015:EMNLP2} and \newcite{YangSC:arxiv16} also used BSLTM to compose character embeddings to word's representation, which is similar to \newcite{lample:2016:NAACL}. \newcite{peng-dredze:2016:ACL} Improved NER for Chinese Social Media with Word Segmentation.

\section{Conclusion}
In this paper, we proposed a neural network architecture for sequence labeling. It is a truly end-to-end model relying on no task-specific resources, feature engineering or data pre-processing. We achieved state-of-the-art performance on two linguistic sequence labeling tasks, comparing with previously state-of-the-art systems.

There are several potential directions for future work. First, our model can be further improved by exploring multi-task learning approaches to combine more useful and correlated information. For example, we can jointly train a neural network model with both the POS and NER tags to improve the intermediate representations learned in our network. Another interesting direction is to apply our model to data from other domains such as social media (Twitter and Weibo). Since our model does not require any domain- or task-specific knowledge, it might be effortless to apply it to these domains.

\section*{Acknowledgements}
This research was supported in part by DARPA grant FA8750-12-2-0342 funded under the DEFT program. 
Any opinions, findings, and conclusions or recommendations expressed in this material are those of the authors and do not necessarily reflect the views of DARPA.

\bibliography{acl2016}
\bibliographystyle{acl2016}

\end{document}